\documentclass{article}

    \PassOptionsToPackage{numbers, compress}{natbib}

    \usepackage[final]{neurips_2020}

\usepackage[utf8]{inputenc} 
\usepackage[T1]{fontenc}    
\usepackage{hyperref}       
\usepackage{url}            
\usepackage{booktabs}       
\usepackage{amsfonts}       
\usepackage{nicefrac}       
\usepackage{microtype}      
\usepackage{amsmath}

\usepackage{algorithm}
\usepackage{booktabs} 
\usepackage{subcaption}
\usepackage{caption}
\usepackage{xcolor}
\usepackage[noend]{algpseudocode}
\usepackage{wrapfig}
\usepackage{adjustbox}
\usepackage{xcolor}

\definecolor{forestgreen}{rgb}{0.14, 0.55, 0.14}

\renewcommand{\paragraph}[1]{\vspace{0.20ex}\noindent\textbf{#1}}

\def\eg{\emph{e.g.~}}

\def\ie{\emph{i.e.~}}

\title{CNNs and Transformers Perceive Hybrid Images Similar to Humans}

\author{Ali Borji \\
Quintic AI, San Francisco, CA \\
\texttt{aliborji@gmail.com} }

\begin{document}

\maketitle

\begin{abstract}

Hybrid images is a technique to generate images
with two interpretations that change as a function of viewing
distance. It has been utilized to study multiscale processing of
images by the human visual system. Using 63,000 hybrid images across 10 fruit categories, here we show that predictions of deep learning vision models qualitatively matches with the human perception of these images. Our results provide yet another evidence in support of the hypothesis that Convolutional Neural Networks (CNNs) and Transformers are good at modeling the feedforward sweep of information in the ventral stream of visual cortex. Code and data is available at \url{https://github.com/aliborji/hybrid_images.git}.

\end{abstract}

\section{Introduction}

\begin{figure}[htbp]
    \centering
    \includegraphics[width=.65\textwidth]{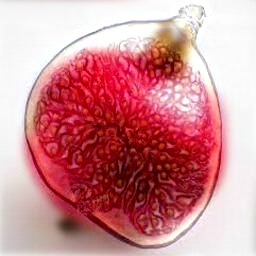} 
    \caption{A sample hybrid image blended from an image of a pomegranate with an image of a fig. What do you see? Step back a few meters and look again! What do you see now? More examples are given in the appendix.}
\label{fig:fig0}
\end{figure}

Hybrid images are a form of illusion which exploits the multiscale visual perspective of human
vision~\cite{oliva2006hybrid}. They are generated by superimposing two images at two
different spatial scales\footnote{Notice that this approach differs from simple interpolation of two images (as is done in \eg mixup~\cite{zhang2017mixup}).}. The low-spatial scale is obtained by filtering one image with a low-pass filter. The high spatial scale is obtained
by filtering a second image with a high-pass filter. The final image is composed by adding these two filtered images. The perception of a hybrid image is a function of viewing distance. When the image is viewed from close distance, it will capture the prominent version of the image and when it is viewed from far distance, it will capture the low-pass version of the image, \ie, the blurry image. 
An example is shown in Fig.~\ref{fig:fig0}.

Our main objective here is to study how state of the art deep vision models perform on hybrid images and whether their predictions matches (at least qualitatively) with our judgement of these images.

\section{Experiments and Results}

\begin{figure}[t]
    \centering
    \includegraphics[width=.09\textwidth]{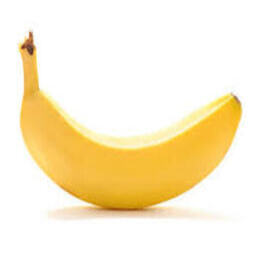}
    \includegraphics[width=.09\textwidth]{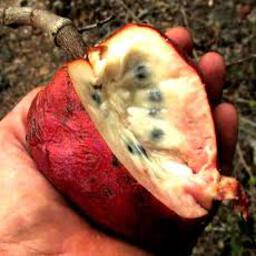}
    \includegraphics[width=.09\textwidth]{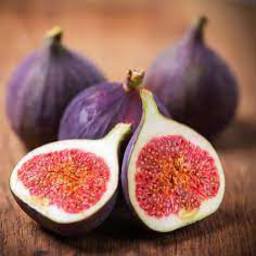}
    \includegraphics[width=.09\textwidth]{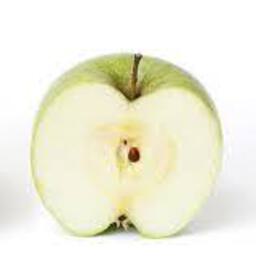}    
    \includegraphics[width=.09\textwidth]{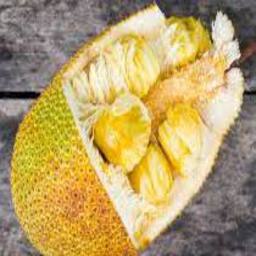}
    \includegraphics[width=.09\textwidth]{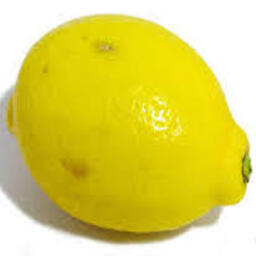}
    \includegraphics[width=.09\textwidth]{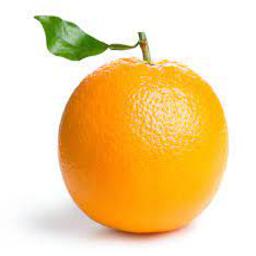}    
    \includegraphics[width=.09\textwidth]{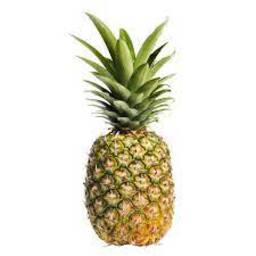}    
    \includegraphics[width=.09\textwidth]{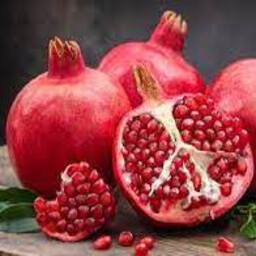}
    \includegraphics[width=.09\textwidth]{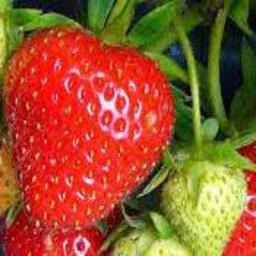}    
    \includegraphics[width=.09\textwidth]{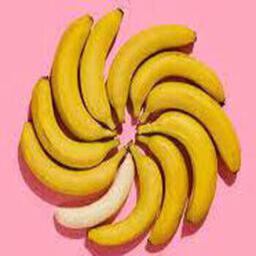}
    \includegraphics[width=.09\textwidth]{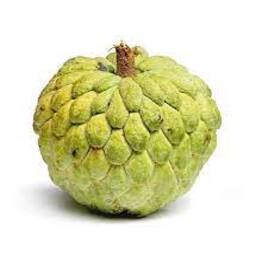}
    \includegraphics[width=.09\textwidth]{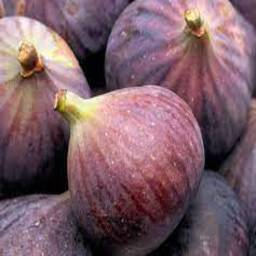}
    \includegraphics[width=.09\textwidth]{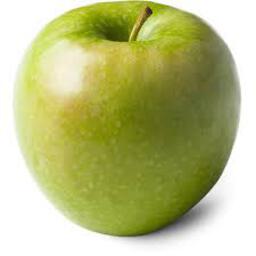}    
    \includegraphics[width=.09\textwidth]{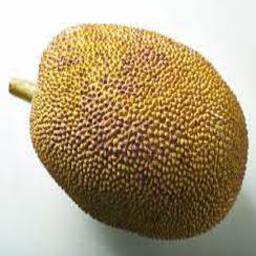}
    \includegraphics[width=.09\textwidth]{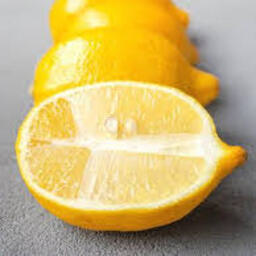}
    \includegraphics[width=.09\textwidth]{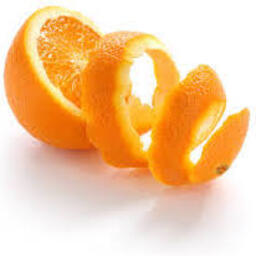}    
    \includegraphics[width=.09\textwidth]{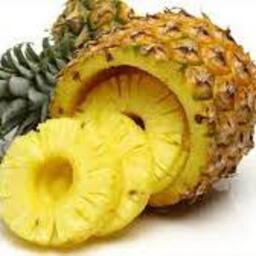}    
    \includegraphics[width=.09\textwidth]{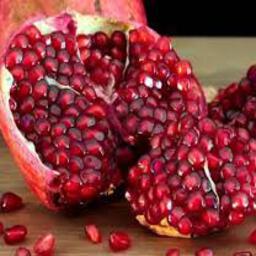}
    \includegraphics[width=.09\textwidth]{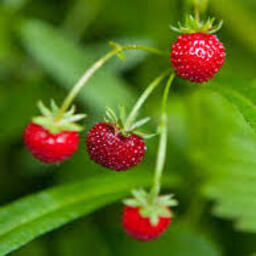}    
    
    \caption{Samples from the 10 fruit categories used in the experiments. We have 10 images per category making it 100 images in total. Categories in order are Banana, Custard Apple, Fig, Granny Smith, Jackfruit, Lemon, Orange, Pineapple, Pomegranate, and Strawberry.}
    \label{fig:Samples}
\end{figure}

We selected 100 images, 10 from each of 10 fruit categories including Banana, Custard Apple, Fig, Granny Smith, Jackfruit, Lemon, Orange, Pineapple, Pomegranate, and Strawberry (images were randomly picked using web search). All of these categories exist in the ImageNet dataset~\cite{deng2009imagenet}, therefore pretrained models on ImageNet should be able to recognize images from these categories. Samples from the dataset are shown in Fig.~\ref{fig:Samples}.

\begin{figure}[t]
    \centering
    \includegraphics[width=.13\textwidth]{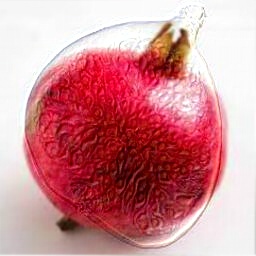}
    \includegraphics[width=.13\textwidth]{Samples/4_output.jpg}    
    \includegraphics[width=.13\textwidth]{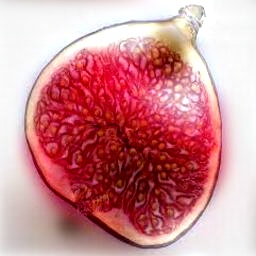}
    \includegraphics[width=.13\textwidth]{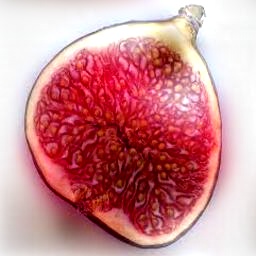}
    \includegraphics[width=.13\textwidth]{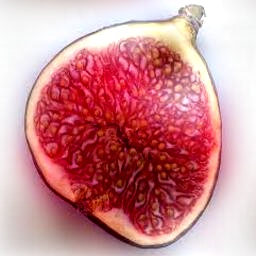}    
    \includegraphics[width=.13\textwidth]{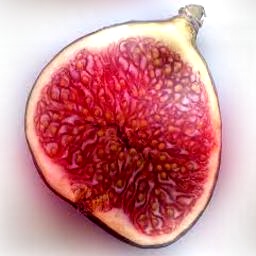}
    \includegraphics[width=.13\textwidth]{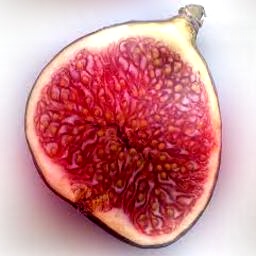}
    \includegraphics[width=.13\textwidth]{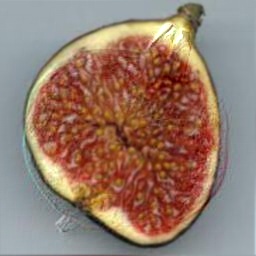}
    \includegraphics[width=.13\textwidth]{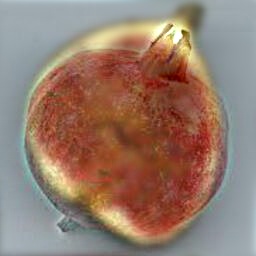}    
    \includegraphics[width=.13\textwidth]{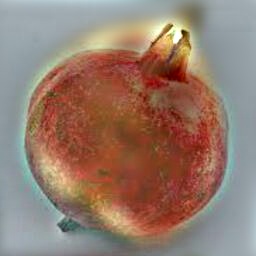}
    \includegraphics[width=.13\textwidth]{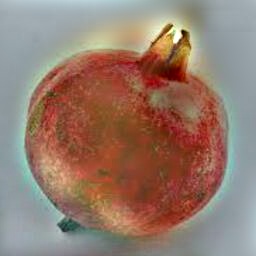}
    \includegraphics[width=.13\textwidth]{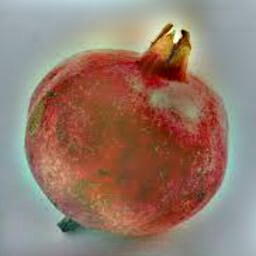}    
    \includegraphics[width=.13\textwidth]{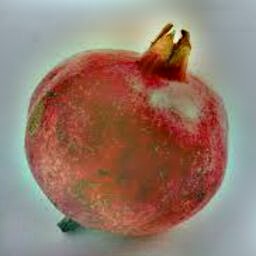}
    \includegraphics[width=.13\textwidth]{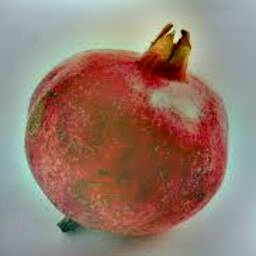}
    
    \caption{Demonstration of hybrid image generation. A high frequency filtered image is added to a low pass filtered image. Columns correspond to cutoff frequencies 1, 4, 7, 10, 13, 16 and 19.}
    \label{fig:teaser}
\end{figure}

We then blended (using the hybrid images technique) all the 10 images in one category to 10 images in another category at 7 different cutoff frequencies (see Fig.~\ref{fig:teaser}), thereby generating $100 \times 7 = 700$ images. Repeating this process for all 90 pairs of categories results in $90 \times 700 = 63000$ images. There are 6300 images mapping images from one category to all other categories. 

We then fed the hybrid images to 6 deep models including 5 CNNs (GoogleNet~\cite{szegedy2015going}, AlexNet~\cite{krizhevsky2012imagenet}, VGG16~\cite{simonyan2014very}, ResNet18 and ResNet50~\cite{he2016deep}), as well as a transformer model (DeiT~\cite{touvron2021training}). All of these models have been pretrained on ImageNet and are available via PyTorch model repository (\url{https://pytorch.org/vision/stable/models.html}).

\begin{figure}[htbp]
    \centering
    \includegraphics[width=.45\textwidth]{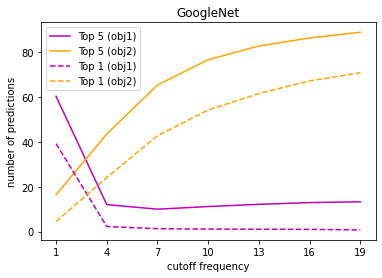}
    \includegraphics[width=.45\textwidth]{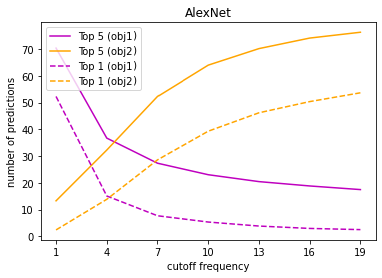} \\
    \includegraphics[width=.45\textwidth]{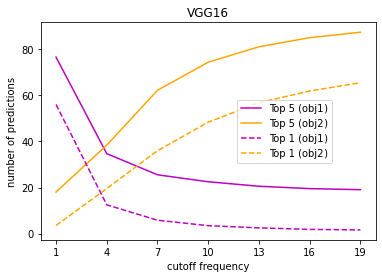} 
    \includegraphics[width=.45\textwidth]{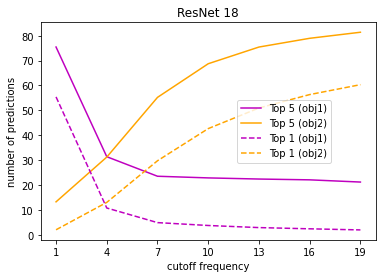} \\
    \includegraphics[width=.45\textwidth]{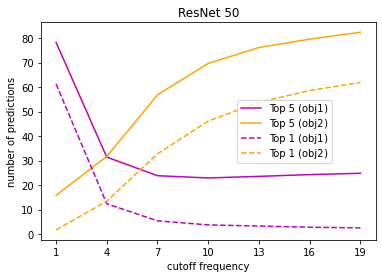}
    \includegraphics[width=.45\textwidth]{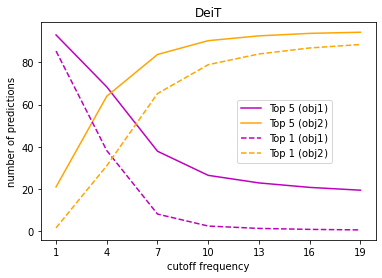} \\
    
    \caption{Performance of deep models on our hybrid images dataset. The $y$ axis shows the number of times (the same as percentage) each blended object appears in the top 1 (or top 5) predictions. As it can be seen, obj1 is reported more at lower cutoff frequencies while at higher cutoff frequencies obj2 is reported more often. Notice that the two curves cross at some point. }
    \label{fig:SUM}
\end{figure}

Results are shown in Fig.~\ref{fig:SUM}. As it can be seen, the low-pass filtered image is reported more often at lower cutoff frequencies, whereas the opposite happens with the high-pass filtered image. Notice that the two curves cross at some point. The model is least certain at the cross point. Interestingly, due to construction of our stimuli, it seems that human perception switches earlier at low cutoff frequencies which also matches with the behavior of models. The cross pattern happens in almost all of category pairs with few exceptions (\eg Orange + Lemon; See Appendix). Among the models, the transformer model outperforms the CNNs.

\section{Discussion and Conclusion}

In this short report, we showed that deep vision models qualitatively behave similar to humans in perceiving the hybrid images. Results suggest that, similar to humans, the deep models process images in a multi-scale fashion where an initial analysis of the global structure is followed by the analysis of local details. Hybrid images paradigm can be used to expand our understanding of deep neural networks. Further, this technique can be used for data augmentation similar to MixUp~\cite{zhang2017mixup}, as well as creating adversarial examples to fool deep models. Finally, extending our work, it would be interesting to a) see if deep models can also be applied to hybrid images consisting more than two images, and b) to inspect the response of different network layers to different frequency components.

\bibliographystyle{plain}
\bibliography{refs}

\clearpage

\appendix
\section{Appendix}

\begin{figure}[t]
    \centering
    \includegraphics[width=1.7\textwidth, height = .7\textheight, angle=90]{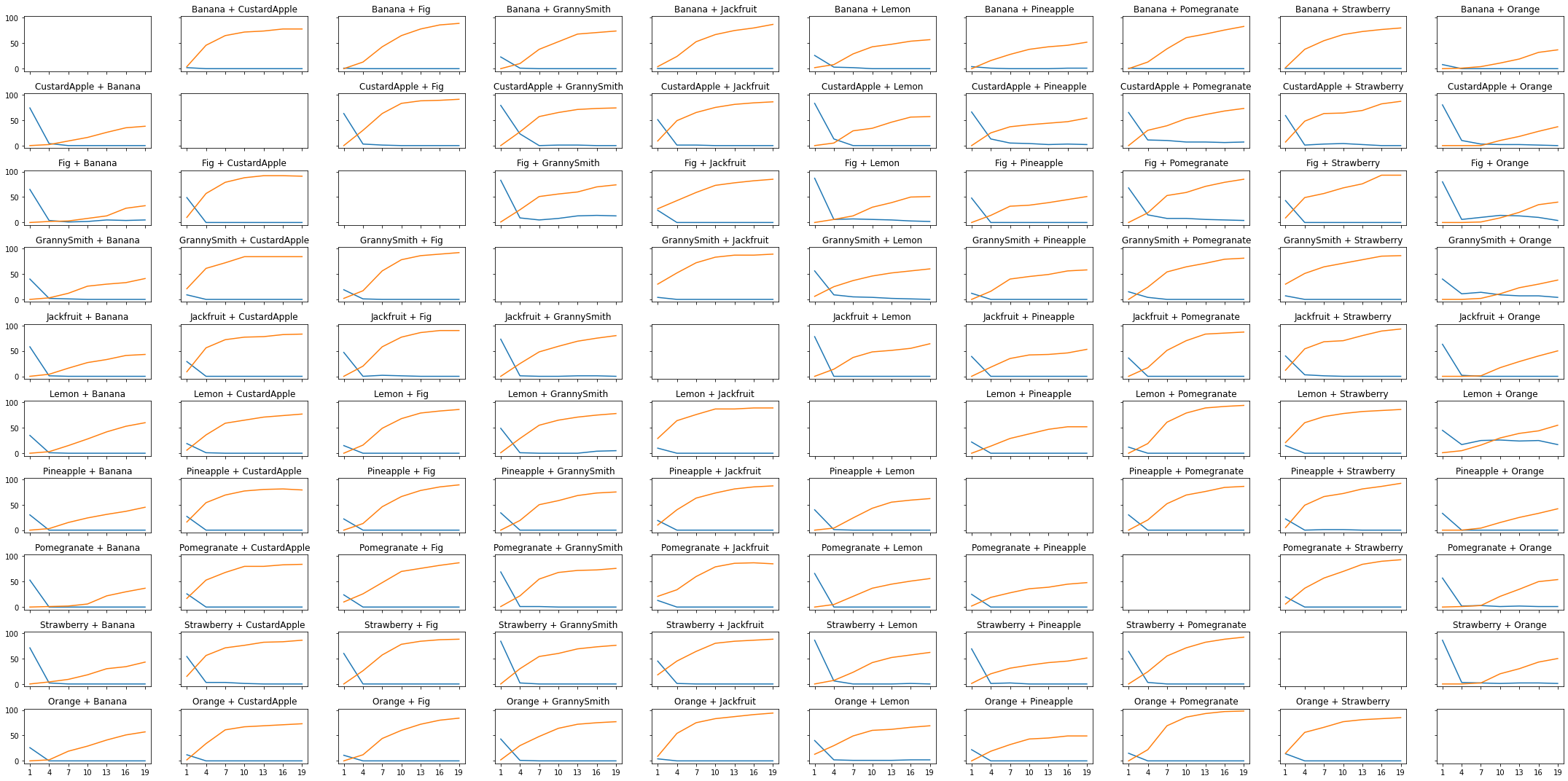}
    \caption{Results over pairs of object categories for GoogleNet (top 1 accuracy).}
    \label{fig:googlenet}
\end{figure}

\begin{figure}[t]
    \centering
    \includegraphics[width=1.7\textwidth, height = .7\textheight, angle=90]{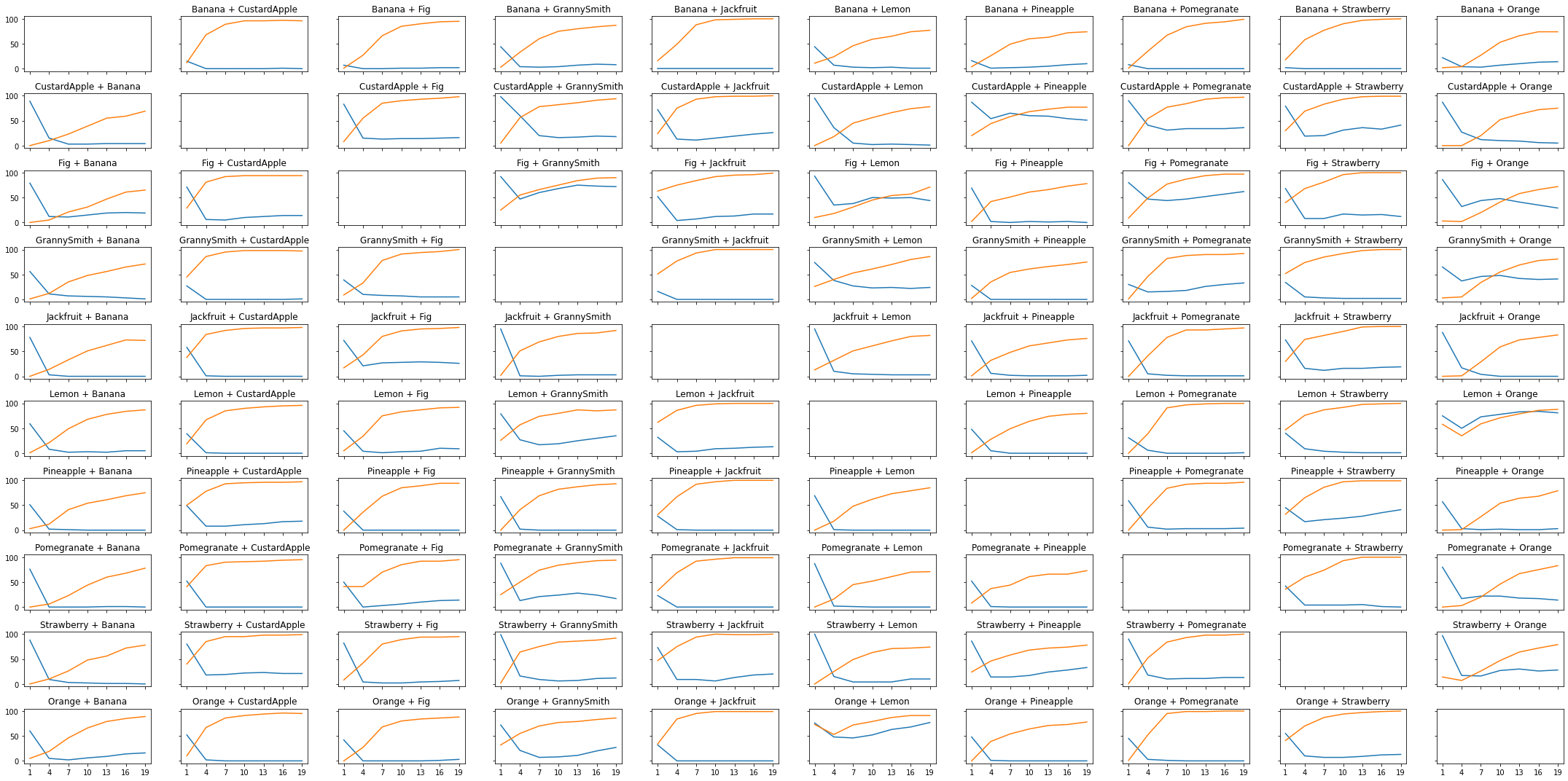}
    \caption{Results over pairs of object categories for GoogleNet (top 5 accuracy).}
    \label{fig:googlenet_t5}
\end{figure}

\begin{figure}[t]
    \centering
    \includegraphics[width=1.7\textwidth, height = .7\textheight, angle=90]{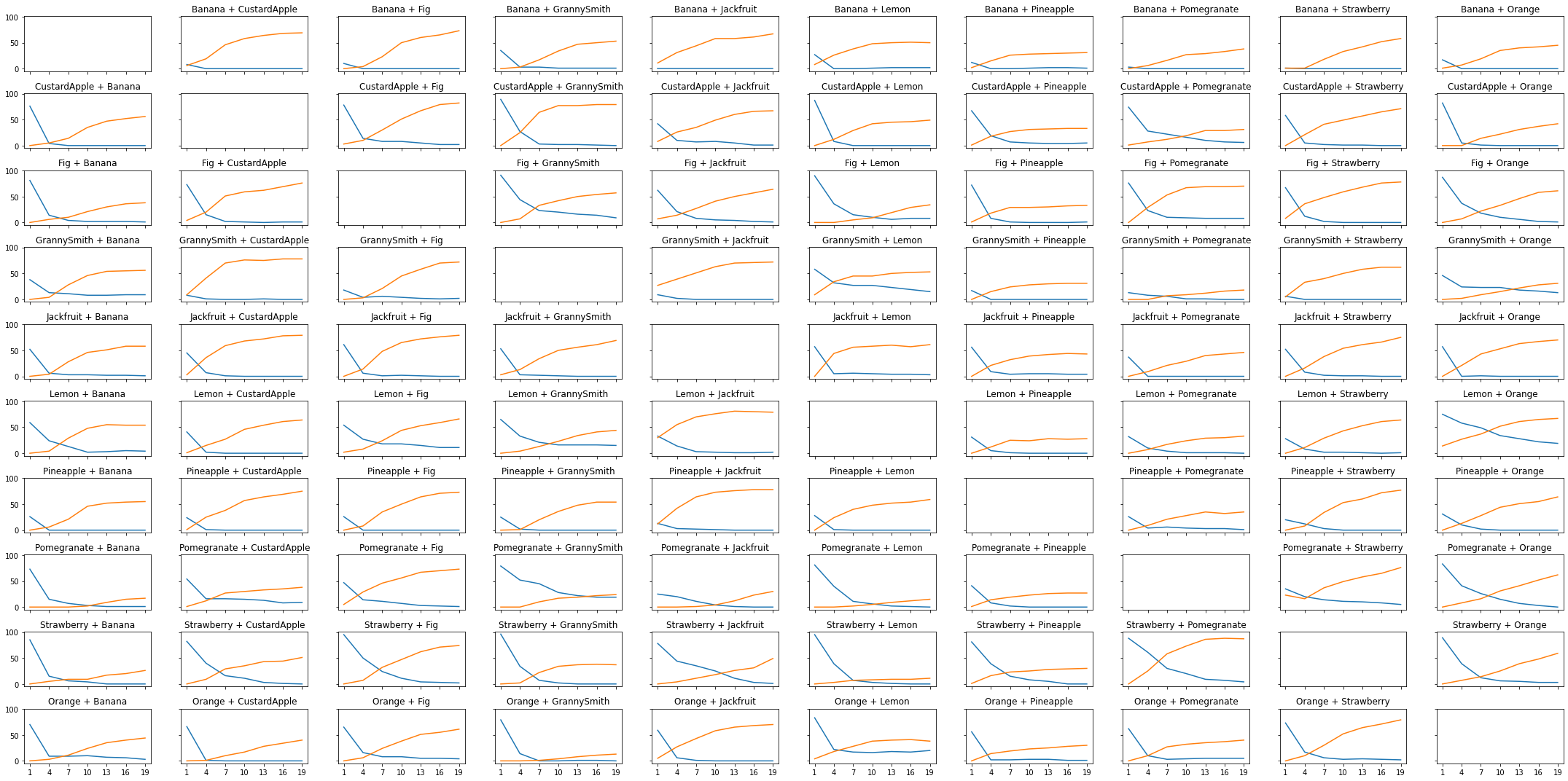}
    \caption{Results over pairs of object categories for AlexNet (top 1 accuracy).}
    \label{fig:alexnet}
\end{figure}

\begin{figure}[t]
    \centering
    \includegraphics[width=1.7\textwidth, height = .7\textheight, angle=90]{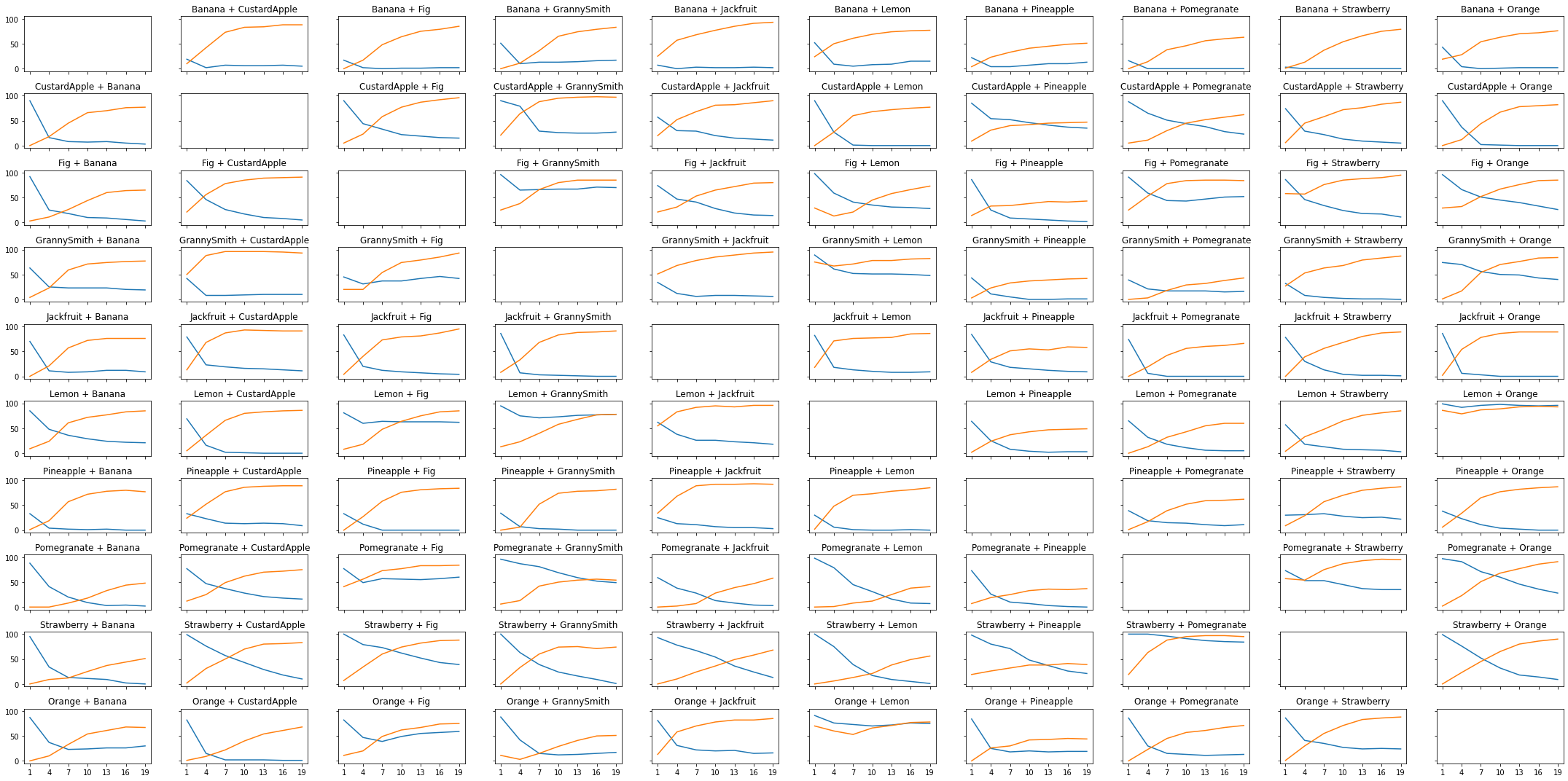}
    \caption{Results over pairs of object categories for AlexNet (top 5 accuracy).}
    \label{fig:alexnet_t5}
\end{figure}

\begin{figure}[t]
    \centering
    \includegraphics[width=1.7\textwidth, height = .7\textheight, angle=90]{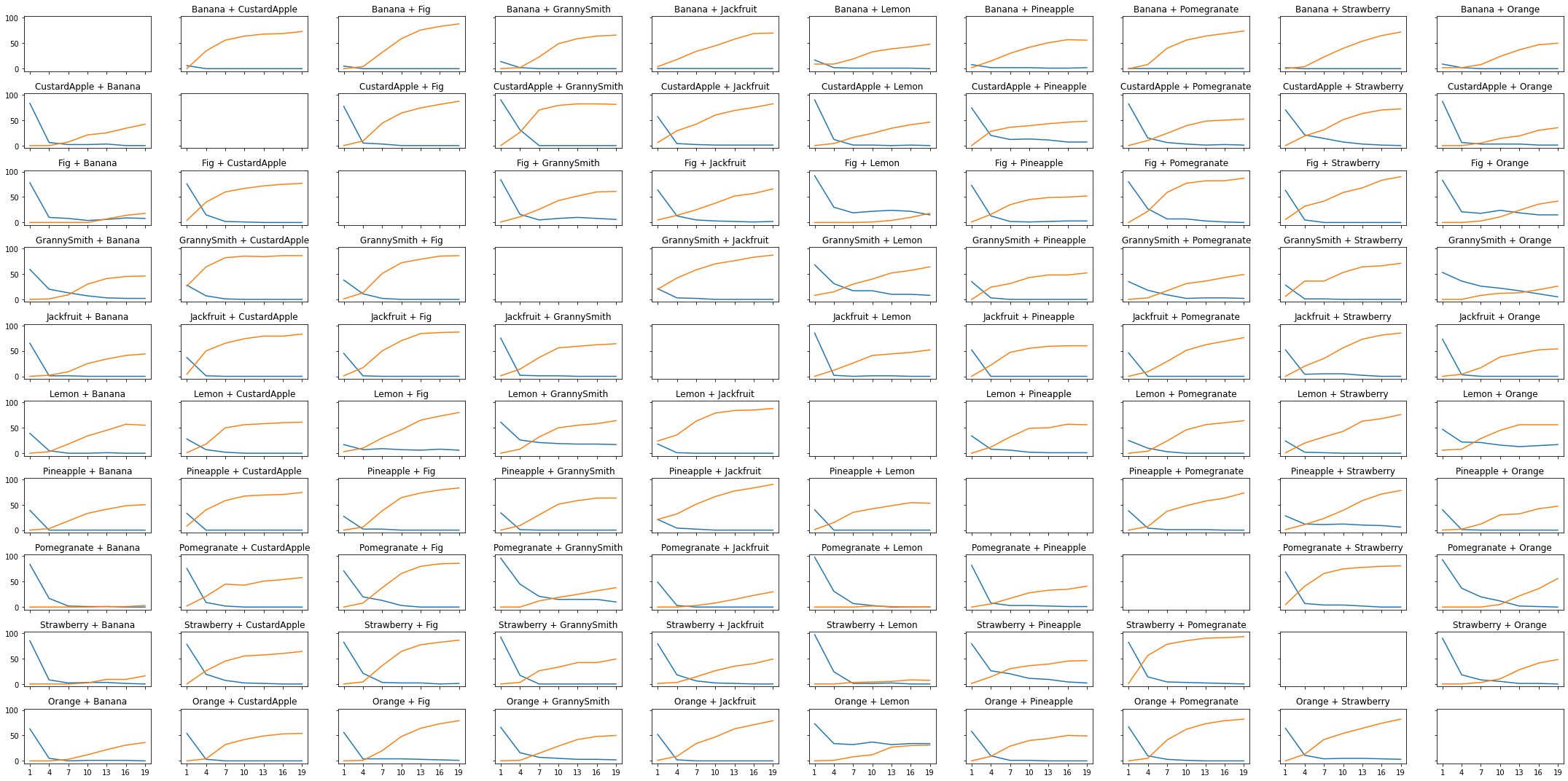}
    \caption{Results over pairs of object categories for ResNet18 (top 1 accuracy).}
    \label{fig:resnet18}
\end{figure}

\begin{figure}[t]
    \centering
    \includegraphics[width=1.7\textwidth, height = .7\textheight, angle=90]{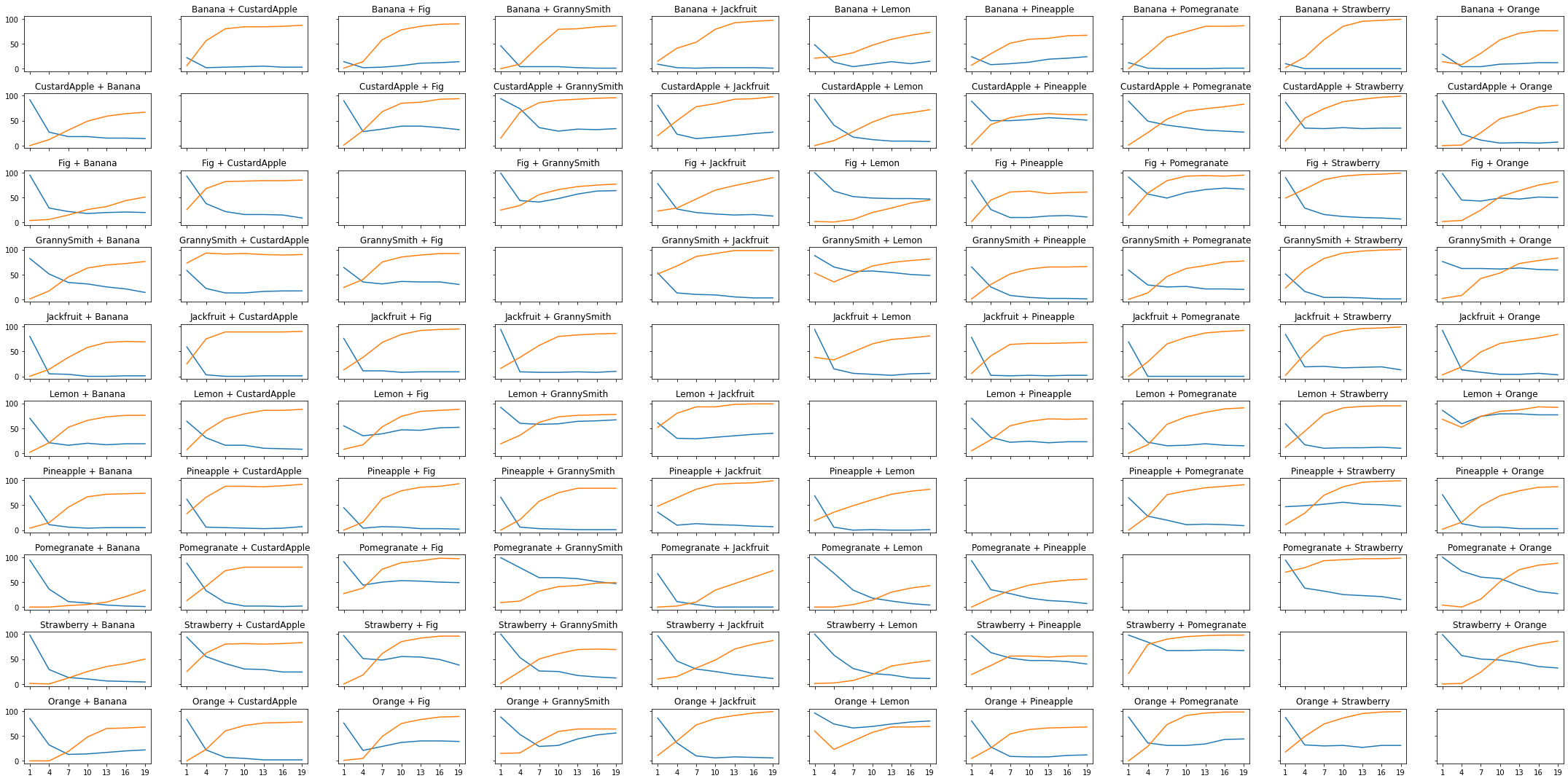}
    \caption{Results over pairs of object categories for ResNet18 (top 5 accuracy).}
    \label{fig:resnet18_t5}
\end{figure}

\begin{figure}[t]
    \centering
    \includegraphics[width=1.7\textwidth, height = .7\textheight, angle=90]{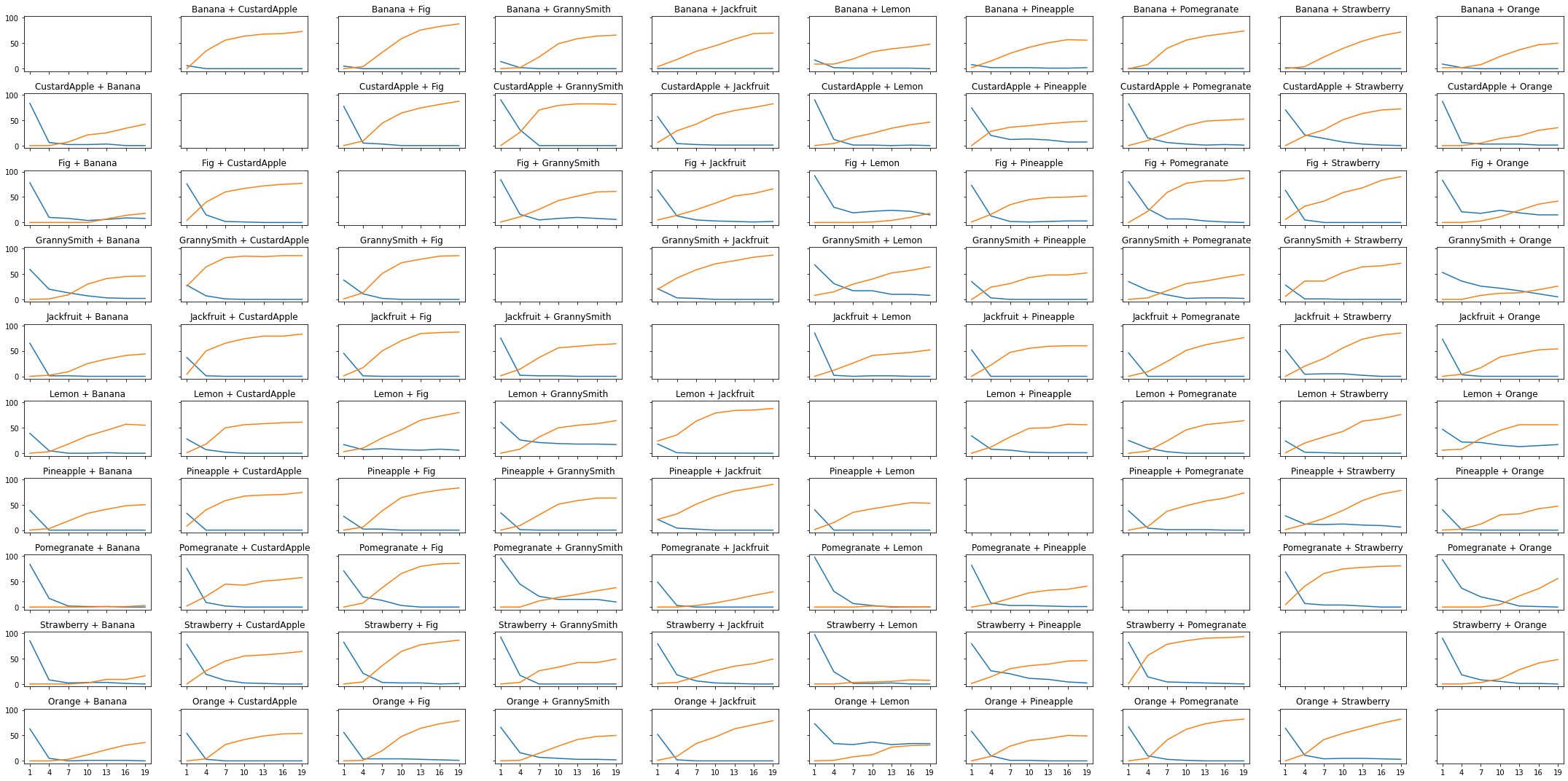}
    \caption{Results over pairs of object categories for ResNet50 (top 1 accuracy).}
    \label{fig:resnet50}
\end{figure}

\begin{figure}[t]
    \centering
    \includegraphics[width=1.7\textwidth, height = .7\textheight, angle=90]{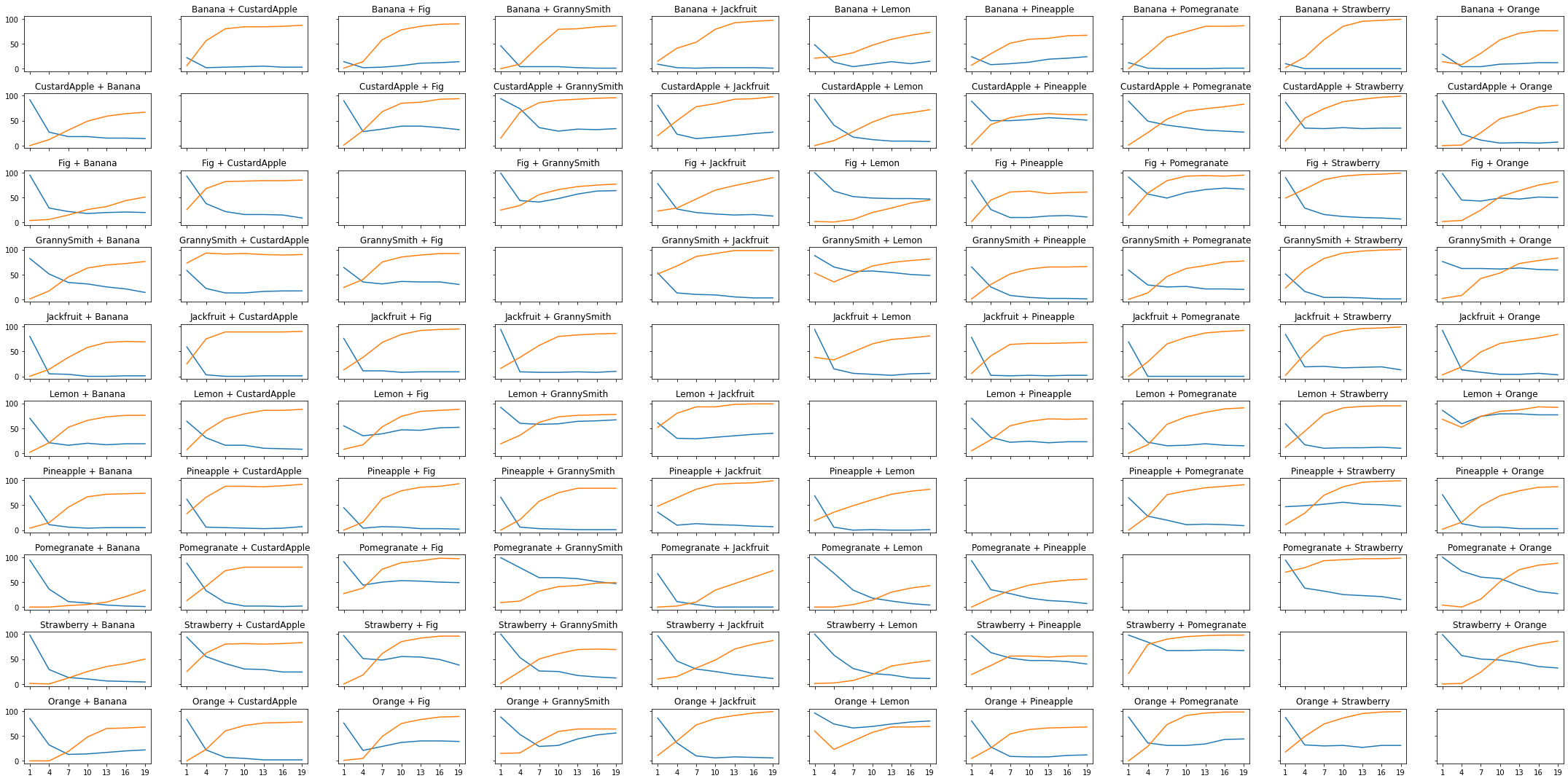}
    \caption{Results over pairs of object categories for ResNet50 (top 5 accuracy).}
    \label{fig:resnet50_t5}
\end{figure}

\begin{figure}[t]
    \centering
    \includegraphics[width=1.7\textwidth, height = .7\textheight, angle=90]{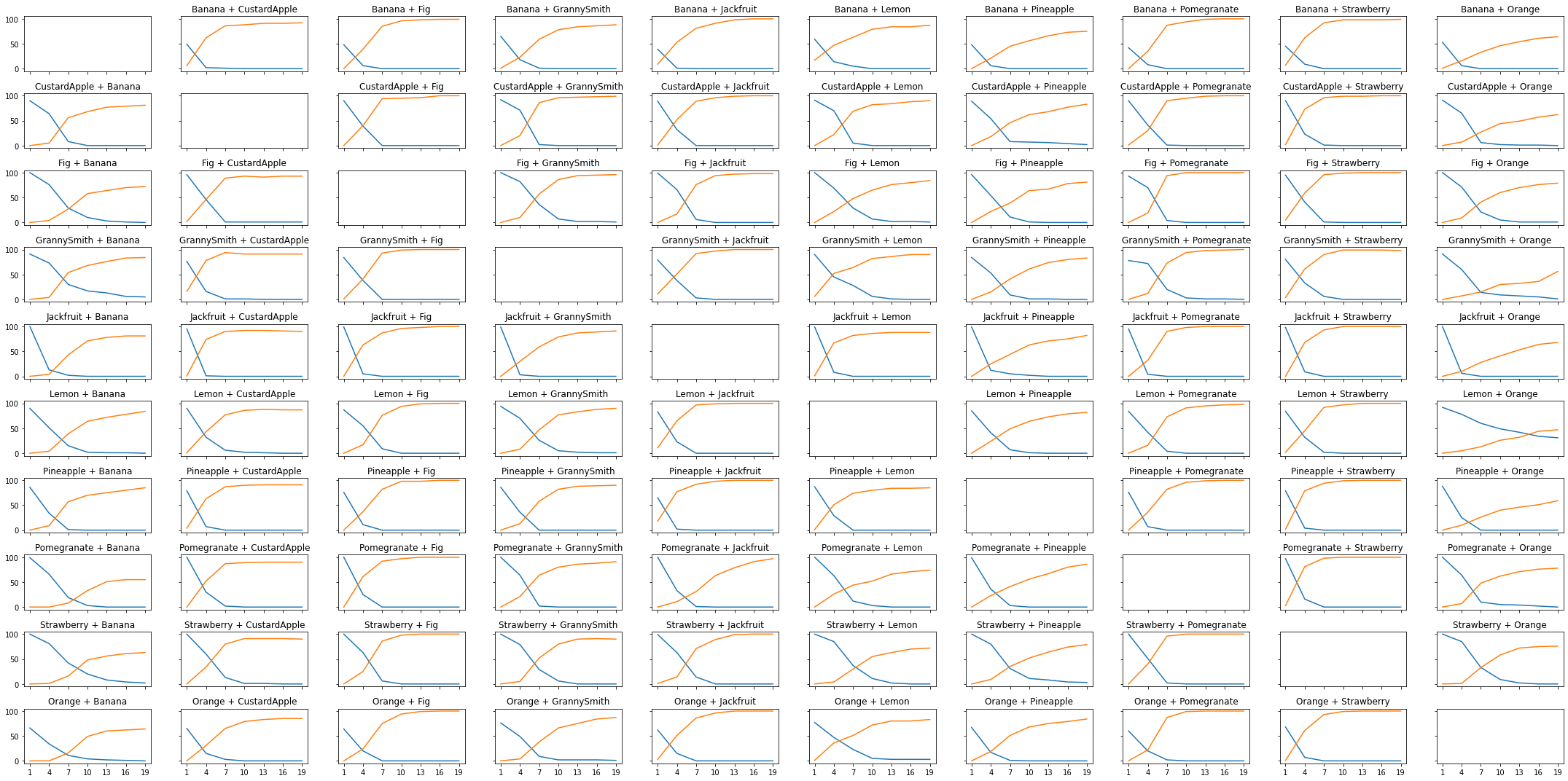}
    \caption{Results over pairs of object categories for DeiT (top 1 accuracy).}
    \label{fig:Dei50}
\end{figure}

\begin{figure}[t]
    \centering
    \includegraphics[width=1.7\textwidth, height = .7\textheight, angle=90]{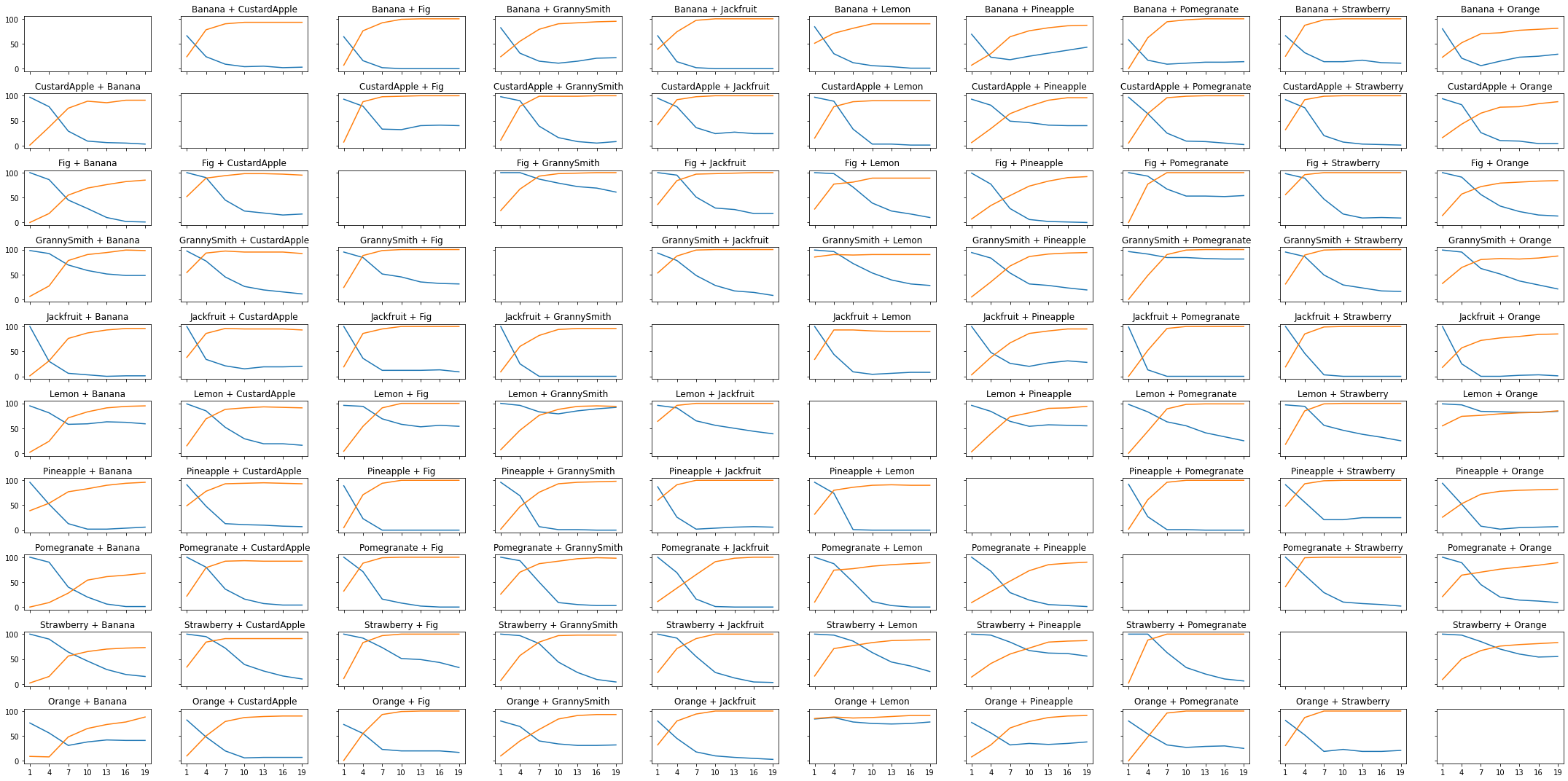}
    \caption{Results over pairs of object categories for DeiT (top 5 accuracy).}
    \label{fig:DeiT_t5}
\end{figure}

\begin{figure}[t]
    \centering
    \includegraphics[width=1\textwidth]{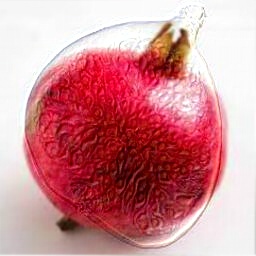}
\end{figure}
    
\begin{figure}[t]
    \centering
    \includegraphics[width=1\textwidth]{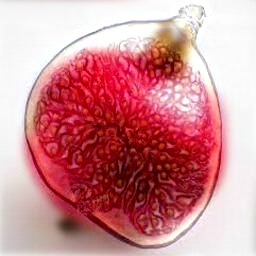} 
\end{figure}

\begin{figure}[t]
    \centering
    \includegraphics[width=1\textwidth]{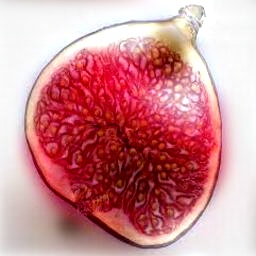}
\end{figure}

\begin{figure}[t]
    \centering
    \includegraphics[width=1\textwidth]{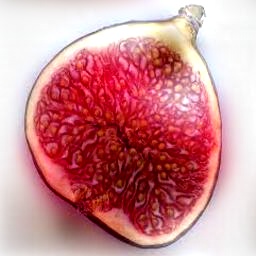}
\end{figure}    

\begin{figure}[t]
    \centering
    \includegraphics[width=1\textwidth]{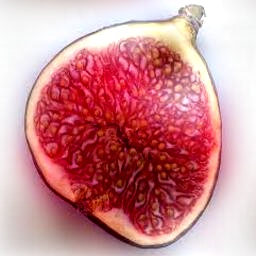}    
\end{figure}    

\begin{figure}[t]
    \centering
    \includegraphics[width=1\textwidth]{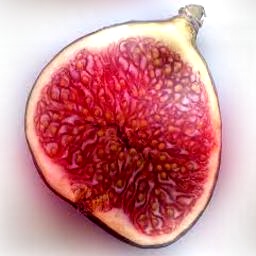}
\end{figure}    

\begin{figure}[t]
    \centering
    \includegraphics[width=1\textwidth]{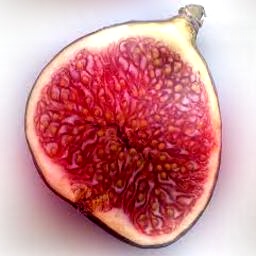}
\end{figure}    

\begin{figure}[t]
    \centering
    \includegraphics[width=1\textwidth]{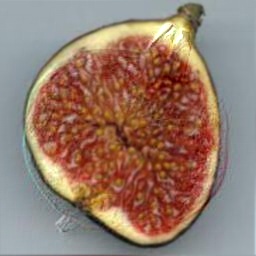}
\end{figure}

\begin{figure}[t]
    \centering
    \includegraphics[width=1\textwidth]{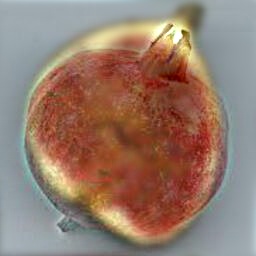}    
\end{figure}    

\begin{figure}[t]
    \centering
    \includegraphics[width=1\textwidth]{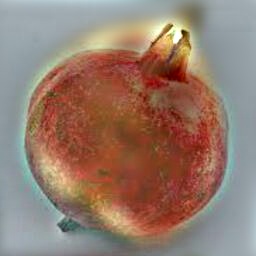}
\end{figure}    

\begin{figure}[t]
    \centering
    \includegraphics[width=1\textwidth]{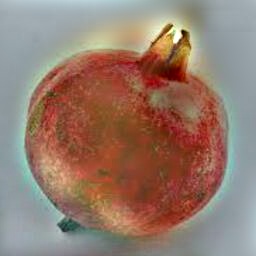}
\end{figure}    
\begin{figure}[t]
    \centering
    \includegraphics[width=1\textwidth]{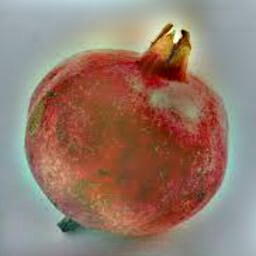}    
\end{figure}    
\begin{figure}[t]
    \centering
    \includegraphics[width=1\textwidth]{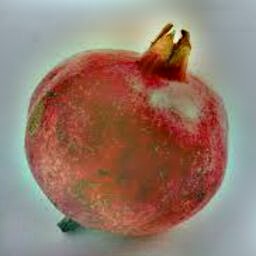}
\end{figure}    
\begin{figure}[t]
    \centering
    \includegraphics[width=1\textwidth]{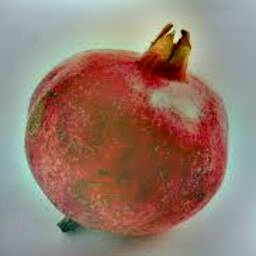}
\end{figure}

\end{document}